\documentclass{article}

     \usepackage[nonatbib, preprint]{neurips_2020}

\usepackage[utf8]{inputenc} 
\usepackage[T1]{fontenc}    
\usepackage{hyperref}       
\usepackage{url}            
\usepackage{booktabs}       
\usepackage{amsfonts}       
\usepackage{nicefrac}       
\usepackage{microtype}      

\usepackage[pdftex]{graphicx}
\usepackage{enumitem,kantlipsum}
\usepackage{caption}
\usepackage{subcaption}

\title{Efficient Incorporation of Multiple Latency Targets in the Once-For-All Network}

\author{
    Vidhur Kumar \\
    Georgia Institute of Technology \\ 
    \texttt{vkumar304@gatech.edu} \\
    \And
    Andrew Szidon \\
    Georgia Institute of Technology \\
    \texttt{aszidon3@gatech.edu}
}

\begin{document}

\maketitle

\begin{abstract}
Neural Architecture Search has proven an effective method of automating \textit{architecture engineering}. Recent work in the field has been to look for architectures subject to multiple objectives such as accuracy and latency to efficiently deploy them on different target hardware. Once-for-All (OFA) is one such method that decouples training and search and is able to find high-performance networks for different latency constraints. However, the search phase is inefficient at incorporating multiple latency targets. In this paper, we introduce two strategies (\textit{Top-down} and \textit{Bottom-up}) that use warm starting and randomized network pruning for the efficient incorporation of multiple latency targets in the OFA network. We evaluate these strategies against the current OFA implementation and demonstrate that our strategies offer significant running time performance gains while not sacrificing the accuracy of the subnetworks that were found for each latency target. We further demonstrate that these performance gains are generalized to every design space used by the OFA network. The code is available \href{https://github.com/vidhur2k/CS8803-Project}{here}.
\end{abstract}

\section{Introduction and Motivation}
Neural Architecture Search (NAS) has proven to be an effective method of automating \textit{architecture engineering}, which is the process of learning a network topology that achieves optimal performance on a particular task. There are three components involved in NAS: a \textit{search space}, which defines the set of all architectures that can be represented in principle, the \textit{search strategy}, which details how the search space should be explored, and the \textit{performance estimation strategy} which helps evaluate candidate architectures and produces feedback for the search algorithm to learn \cite{nassurvey}. Recent advances in NAS has moved beyond solely looking for architectures with high accuracy by also aiming to efficiently deploy these models on \textit{diverse} hardware platforms given certain resource constraints (latency, FLOPs, energy, etc). This is a more difficult problem to solve because considering multiple objectives adds more complexity to an already combinatorial design space. 

Many different approaches have been proposed to efficiently find architectures that meet a certain accuracy-latency tradeoff. Of these, \textit{Once-For-All} (OFA) is a method that consistently outperforms the state-of-the-art on diverse hardware platforms. The idea behind OFA is to train a once-for-all network that supports diverse architectural settings by decoupling the training and search process. The training process uses a \textit{progressive shrinking} algorithm to fine-tune the largest network to support smaller sub-networks. The search process involves using a prebuilt latency lookup table for a target hardware along with an accuracy predictor (a DNN) and running an \textit{evolutionary search} to obtain a specialized sub-network that meets the accuracy-latency tradeoff \cite{ofa}. Given that OFA was created to support diverse hardware platforms (each with varying latency constraints), a natural question that arises is whether or not OFA \textit{efficiently} incorporates multiple latency targets, searches for and outputs the corresponding family of optimal architectures. If we consider 10 latency targets with OFA's current implementation, finding the optimal architecture for each target would take a few minutes. While that does not seem a time-consuming process, consider these scenarios where we deal with \textit{dynamic hardware:}
\begin{enumerate}[leftmargin=*]
    \item \textbf{Datacenters:} Data centers often upgrade a part of their hardware to support emerging workloads. Most hardware is retained for 4 years on average, which is relatively long time considering the speed at which hardware is growing \cite{datacenters}. Each datacenter is thus likely to house a multitude of different hardware. In context of OFA, each of these correspond to a different latency constraint that demands searching for a subnetwork that meets said latency constraint. 
    \item \textbf{Android OS Phones:} Since Android OS is an open-source system, a lot of phones \textit{with different combinations of hardware components} (Processor, RAM, etc.) have been released and are in use by the public. Each phone corresponds to a different \textit{target hardware}. If we wish to run a DNN for a particular task on each of these phones, we need to find a DNN that meets a particular latency constraint for each target hardware. In context of OFA, we need to search for a subnetwork that meets the latency constraints of each target hardware and output them accordingly.
\end{enumerate}

Therefore, dealing with dynamic hardware includes regular searching of the optimal sub-networks for each latency constraint. OFA's current implementation suffers from having to search for sub-networks meeting different latency constraints \textit{separately}. Algorithmically speaking, if we were dealing with $k$ latency constraints where the population was evolved for $n$ iterations for each evolutionary search we have a running time of $\mathcal{O}(kn)$.


We introduce two new strategies for addressing the aforementioned inefficiency: 1. A \textit{top-down} search that considers the latency constraints in descending order and \textit{warm starts} the evolutionary search for a particular latency constraint with a \textit{randomly pruned version} of the previous latency constraint. 2. A \textit{bottom-up} search that considers the latency constraints in ascending order and \textit{warm starts} the evolutionary search for a particular latency constraint with the optimal sub-network of the previous latency constraint. The rest of this paper is divided as follows: Section 2 provides an overview of some related works significant to our contribution. Section 3 provides a detailed explanation of our strategies along with an algorithmic analysis. Section 4 covers a comprehensive set of experiments we ran to evaluate our strategies against the current OFA implementation. Section 5 covers some broader impacts of our work.

\section{Related Work}
\paragraph{Latency-Aware NAS:} Recent work in the field of NAS has focused on optimizing for multiple objectives. One class of NAS algorithms focuses on finding pareto-optimal architectures that meet an accuracy-latency tradeoff. MNasNet is a mobile neural architecture search method that explicitly incorporates real-world inference latency into the main objective so that the search can identify an architecture that achieves an optimal accuracy-latency tradeoff \cite{mnasnet}. Shah and El-Sharkawy introduce two variations of MNasNet named A-MNasnet and R-MNasnet in \cite{a-mnasnet} and \cite{r-mnasnet} respectively. The former introduced a new architecture based on MNasNet that makes the model more efficient in terms of accuracy, and the latter introduces a modified architecture based on MNasNet that makes it more compact with a fair trade-off between model size and accuracy. CompOFA addresses the issue of OFA suffering from a combinatorial explosion of model configurations by incorporating insights of compound relationships between model dimensions and restricting the design space to configurations with better accuracy-latency tradeoffs \cite{compofa}.

\paragraph{Hardware-Efficient Deep Learning:} This subfield of Deep Learning focuses on optimizing the hardware efficiency of high-performance DNNs. SqueezeNet maintained AlexNet-level accuracy with 50x fewer parameters \cite{squeezenet}. MobileNets are streamlined architectures for mobile and embedded vision applications that use depthwise-separable convolutions to build lightweight DNNs that efficiently trade off between latency and accuracy \cite{mobilenet}. "Deep Compression" is a technique that involves a three stage pipeline: pruning, trained quantization, and huffman coding that significantly reduces the storage requirement of DNNs by retaining the most important connections \cite{deepcompression}.

\paragraph{Warm Starting DNN Training:} Training DNNs involves solving a sequence of optimization problems. \textit{Warm Starting} is a technique that initializes the optimization task with the solution of the previous iterate rather than starting from scratch. It is purely intuitive but has seen great success as it relies on the fact that the search for the optimal parameters is anchored to a good starting point \cite{warmstarting}.

\section{Design and Implementation}
\cite{ofa} introduces the OFA strategy where the\textit{training} of the large network and the \textit{search} for a subnetwork that meets a latency constraint are decoupled. Our strategies are targeted towards the \textit{search} phase of this process when there are multiple latency constraints to consider. 
\subsection{Top-down Strategy}
The \textit{Top-down} strategy assumes that we receive $k$ latency targets $T = [T_1, T_2, \dots, T_k]$ in decreasing order ($T_1 > T_2 > \dots > T_k)$ The \textit{top-down} strategy works off the intuition that if we have found an optimal subnetwork for some $T_i$, then the parameters of a 'smaller' version of this subnetwork would be a good starting point for the search of the optimal subnetwork for $T_{i+1}$. Therefore, for each latency target $T_i$ where $i \in [2, k]$ we add a \textit{randomly pruned} version the optimal subnetwork to the population of models being considered during evolutionary search. The number of evolutions is set to $n$ for $T_1$, but for every subsequent latency target, we set the number of evolutions to $\sqrt{n}$ as the search for the optimal subnetwork is already anchored to a good starting point rather than starting from scratch. 

\subsection{Bottom-up Strategy}
The \textit{Bottom-up} strategy assumes that we receive $k$ latency targets $T = [T_1, T_2, \dots, T_k]$ in non-decreasing order ($T_1 < T_2 < \dots < T_k)$. The \textit{bottom-up} strategy works off the intuition that the parameters of the optimal subnetwork for some $T_i$ for $i \in [1, k]$ is a good starting point for the search of the optimal subnetwork for $T_{i+1}$. Therefore, for each latency target $T_i$ where $i \in [2, k]$ we add the optimal subnetwork found for $T_{i-1}$ to the population of models being considered during evolutionary search. Given that this was the fittest model, it would be a good starting point for $T_i$. The number of evolutions is set to $n$ for $T_1$, but for every subsequent latency target, we set the number of evolutions to $\sqrt{n}$ as the search for the optimal subnetwork is already anchored to a good starting point rather than starting from scratch. 

\begin{figure}[!h]
  \centering
  \includegraphics[width=0.8\linewidth]{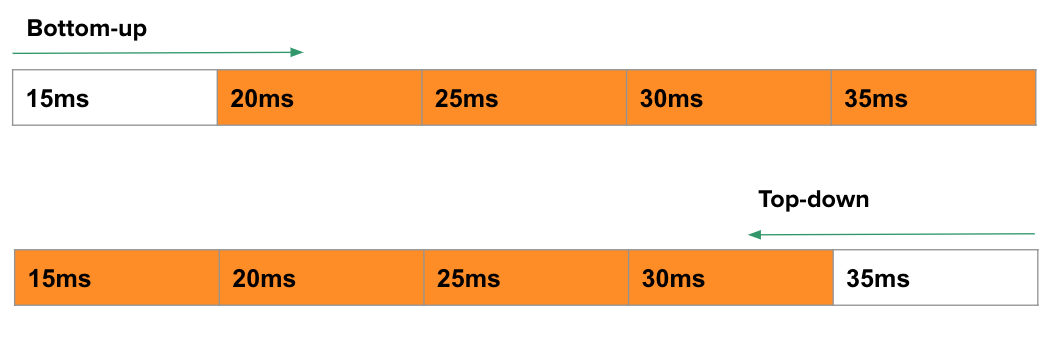}
  
  \caption{An instance of the \textit{top-down} and \textit{bottom-up} strategy being applied for 5 latency targets. Notice that the two strategies iterate in opposite directions. Orange cells represent latency targets that have their evolutionary search \textit{warm started} with the previous latency target's optimal sub-network (randomly pruned sub-network for \textit{top down}).}
\end{figure}

\subsection{Algorithmic Analysis of the Strategies}
For this section, we assume that there are $k$ latency targets and the number of evolutions during the search for the optimal subnetwork of each target is $n$. We also hold all other parameter equal for all strategies. 

\paragraph{Original OFA:} Let $T$ be the running time of the algorithm. Since the original OFA implementation requires that we handle each latency target \textit{separately}, we have:
\begin{center}
    $T = n + n + \dots + n$ ($k$ times) \\
    $T = kn$
\end{center}
Therefore, $T \in \mathcal{O}(kn)$ for the original OFA implementation.
\paragraph{Top-down and Bottom-up}: Let $T'$ be the running time of the algorithm. Notice that we make the same changes to $n$ in both strategies. Hence, it makes sense to perform a single algorithmic analysis for both strategies. Since we reduce the number of evolutions to $\sqrt{n}$ starting from the second latency target, we have:
\begin{center}
    $T' = n + (k-1)\sqrt{n}$ \\
    $T' = n + k\sqrt{n} - \sqrt{n}$
\end{center}
Since the $\sqrt{n}$ term at the end is irrelevant for large values of $n$, we have that $T' \in \mathcal{O}(n + k\sqrt{n})$.

Recall the two scenarios we posed in Section 1 with the datacenters and Android OS phones. In those two contexts, it is reasonable to conclude that we would have a large value of $k$ (somwehere in the hundreds). Assume for the sake of argument that the value of $k \geq n$ (the value of $n$ is typically set to 500 in most cases and there are over 500 types of Android phones or datacenter hardware, which makes the assumption valid). In this case, $T \in \mathcal{O}(kn) \approx \mathcal{O}(n^2)$, whereas $T* \in \mathcal{O}(n + k\sqrt{n}) \approx \mathcal{O}(n + n\sqrt{n}) = \mathcal{O}(n^{3/2})$. Therefore, in the scenario where there is a large number of target hardware to consider, the current OFA implementation runs in quadratic time, while our algorithms is sub-quadratic. This makes our approach the obvious choice in said scenarios.

\subsection{Limitations to our Approach}
\begin{figure}[!h]
    \centering
    \includegraphics[width=0.8\textwidth]{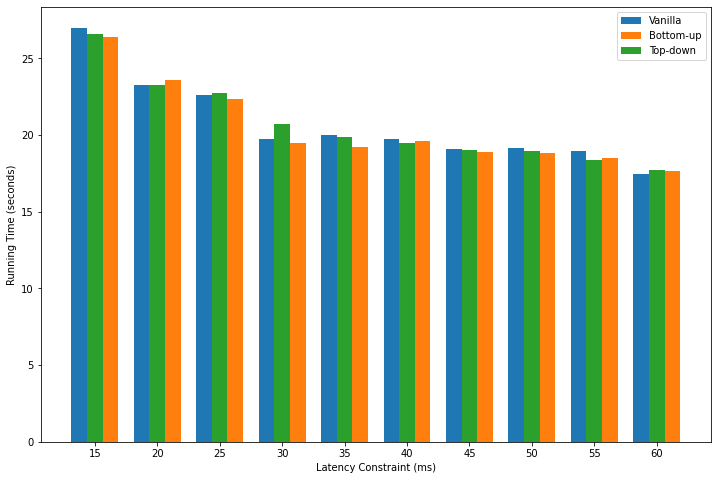}
    \caption{The running time of finding the optimal subnetwork for a single latency constraint using the original OFA method, the \textit{top-down}, and the \textit{bottom-up} strategy}
    \label{fig:single}
\end{figure}

\paragraph{The Need for an Accuracy and Latency Predictor:} While our approach does efficiently incorporate multiple latency targets, it is still dependent upon the accuracy predictor and the latency lookup table that the original OFA method utilizes. It is important to realize that while the accuracy and latency predictor are reasonable estimates of the performance of a particular subnetwork, they are still an estimate at the end of the day. The fact remains that the accuracy and latency values that the search hinges on may not always be accurate, thereby leading to sub-optimal subnetworks being found at times.
\paragraph{The Cost of Finding the Optimal Subnetwork for a \textit{Single} Latency Constraint:} The \textit{Top-down} and \textit{Bottom-up} strategies, while effectively amoritizing the cost of finding subnetworks for multiple latency targets, do not provide any improvement to the cost of finding a subnetwork for a single latency constraint. Algorithmically speaking, this does make sense: our strategies perform $n$ evolutions while searching for the subnetwork corresponding to the \textit{first} latency constraint and then reduce the number of evolutions for the subsequent searches. Naturally, these strategies would perform the same number of evolutions as the original OFA method when there is only one latency target to consider. As shown in Figure \ref{fig:single}, there is not much difference in the running times when there is only one latency target to consider, no matter what the constraint is. Also notice that there \textit{it takes a slightly longer time to find the subnetwork when the latency target is lower}, i.e., there is \textit{negative correlation between the running time of the search and the latency target}. This can be attributed to how this implementation of OFA enforces a hard latency constraint, where whenever an attempt to randomly generate or mutate a candidate subnetwork fails to meet the latency constraint, the algorithm repeats that attempt until a valid subnetwork is found, and the lower the latency constraint, the more likely more repeats are required.

\section{Evaluation}
\subsection{Experimental Setup}
We perform all experiments on an Intel Core i9 2.9GHz (4.8GHz) Hexa-core processor and utilize 16GB DDR3 RAM. For all experiments, the \textit{top-down} and \textit{bottom-up} approach have the number of evolutions $n$ set to $500$ for the first latency target and then $\sqrt{500} \approx 63$ evolutions for every subsequent target. For each experiment, we compare the performance of 3 strategies: the original (or 'vanilla') OFA method, the \textit{top-down} strategy and the \textit{bottom-up} strategy. We perform the following experiments to test the efficacy of our method (\textbf{Note:} The running times and accuracies for each of the experiments is averaged over 10 runs to ensure statistical significance.):
\begin{enumerate}[leftmargin=*]
    \item Test the \textbf{running time} of our approach against the original OFA search for multiple latency targets. We average the running time over 10 runs to ensure statistical significance.
    \item Ensure that the \textit{top down} or \textit{bottom up} search does not lead to a loss in \textbf{accuracy} of the specialized sub-networks. The accuracy is also averaged over 10 runs.
    \item Ensure that the performance gains that we realize with our approach generalize to all \textbf{design spaces} used in OFA (MobileNetV3, ResNet50D, and ProxylessNAS).
\end{enumerate}
\subsection{Experimental Evaluation}
\paragraph{Running Time}
The running time of the algorithm is the primary indicator of the performance gains that our strategies demonstrate. One thing we noted while observing the performances for different numbers of latency targets is that the \textit{difference in performance between the vanilla method and the two strategies becomes more significant with a higher number of latency targets.} This is extremely promising as the motivation behind introducing these strategies was to be able to incorporate a large number of latency targets in the OFA network. As shown in Table 1, the \textit{top-down} method, on average, takes $\textbf{28.3\%}$ of the time that the vanilla method takes, while the \textit{bottom-up} method, on average, takes $\textbf{41\%}$ of the time that the vanilla method takes. The \textit{top-down} method comes out on top for any number of latency targets considered. From a superficial look at both algorithms, it might not seem obvious as to why \textit{top-down} performs slightly better. But, recall the negative correlation between the running time and the latency constraint that we noticed in Figure \ref{fig:single}. Notice that the \textit{top-down} approach starts with the highest latency target while the \textit{bottom-up} approach starts with the lowest latency target. Since both strategies start with $n$ evolutions and cut down to $\sqrt{n}$ evolutions after the first latency target, it makes sense that \textit{top-down} will be slightly faster since it is able to find the optimal subnetwork for its first latency target faster than \textit{bottom-up} is able to for its latency target. Note that this discrepancy may not generalize to a different algorithm that does not share the negative correlation between latency constraint and running time.
\begin{figure}[!h]
    \centering
    \includegraphics[width=0.8\textwidth]{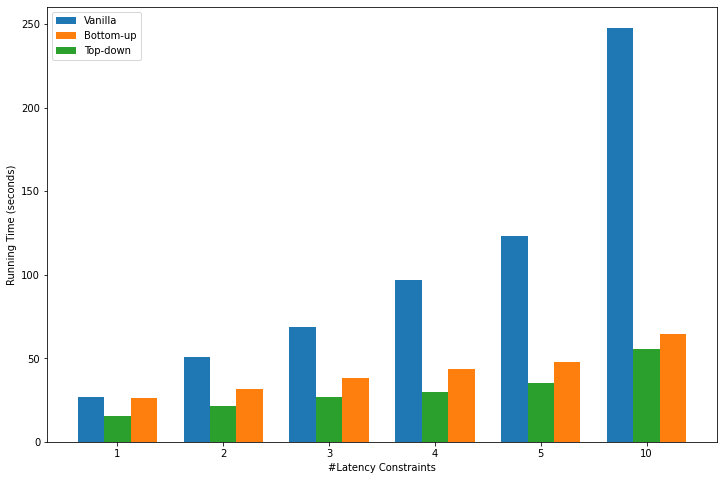}
    \caption{The running time of the 3 strategies visualized in a bar plot. Blue is the vanilla OFA method, orange is bottom-up, and green is top-down.Notice that the difference in running times becomes more significant as the number of latency targets increases. The design space used here is MobileNetV3.}
    \label{fig:mobilenetrunningtime}
\end{figure}

\begin{table}[h]
  \label{tab:mobilenetrunningtime}
  \centering
  \begin{tabular}{llll}
    \toprule \toprule
    \multicolumn{1}{c}{} & \multicolumn{3}{c}{\textbf{Running Time (s)}} \\
    \midrule
    \textbf{\#Latency Targets} & \textbf{Vanilla} & \textbf{Top-down} &
    \textbf{Bottom-up} \\
    \midrule
    1 & $27.065 \pm 0.72$ & \textbf{15.352 $\pm$ 1.21} & $26.166 \pm 0.84$ \\
    2 & $50.516 \pm 1.21$ & \textbf{21.317 $\pm$ 1.04} & $31.591 \pm 1.44$ \\
    3 & $68.795 \pm 1.60$ & \textbf{26.708 $\pm$ 1.47} & $38.235 \pm 1.72$ \\
    4 & $96.670 \pm 2.17$ & \textbf{29.722 $\pm$ 1.80} & $43.809 \pm 2.03$ \\
    5 & $123.109 \pm 2.98$ & \textbf{35.341 $\pm$ 1.83} & $55.608 \pm 2.16$ \\
    10 & $247.858 \pm 5.12$ & \textbf{47.781 $\pm$ 1.91} & $64.291 \pm 2.37$ \\
    \bottomrule
    \bottomrule \\
  \end{tabular}
    \caption{The running time of the 3 strategies used for efficient incorporation of multiple latency targets. The design space used here is MobileNetV3. The fastest performing algorithm is highlighted.}
\end{table}

\paragraph{Accuracy}
We consider the validation accuracies of all the subnetworks found as the \textit{algorithmic} performance gains realized by our two strategies should not lead to a significant loss in the \textit{model} performance. As shown in Table 2, the accuracies of the subnetworks obtained using the 3 strategies show no significant difference at all. The highest difference in accuracy (averaged over 10 runs) we saw was $\textbf{0.58\%}$ for a latency constraint of 50ms, and this was when \textit{top-down} came out on top of the other two. The average difference in accuracy between the lowest and highest accuracy of each latency constraint was $\textbf{0.21}\%$. \begin{figure}[!h]
    \centering
    \includegraphics[width=0.8\textwidth]{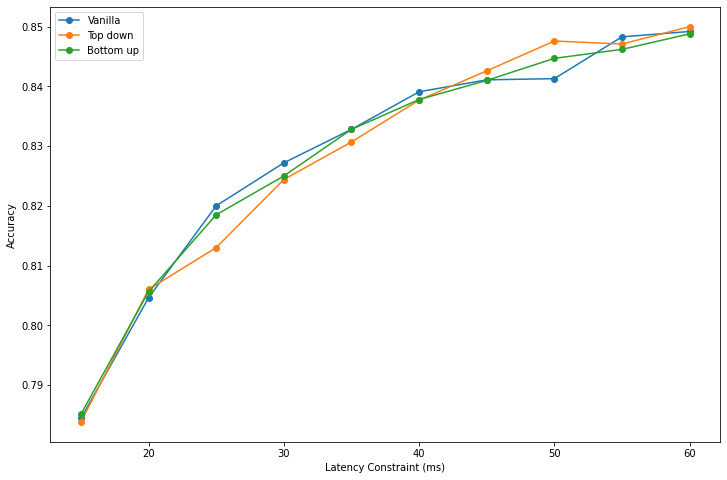}
    \caption{The validation accuracy of the optimal subnetworks found for each latency constraint by the 3 strategies. The design space used here is MobileNetV3.}
    \label{fig:my_label}
\end{figure}

\begin{table}[!h]
  \label{mobileacc}
  \centering
  \begin{tabular}{llll}
    \toprule \toprule
    \multicolumn{1}{c}{} & \multicolumn{3}{c}{\textbf{Accuracy (\%)}} \\
    \midrule
    \textbf{Latency Constraint (ms)} & \textbf{Vanilla} & \textbf{Top-down} & \textbf{Bottom-up} \\
    \midrule
    15 & $0.7844 \pm 0.09$ & $0.7838 \pm 0.12$ & \textbf{0.7851 $\pm$ 0.13} \\
    20 & $0.8046 \pm 0.07$ & \textbf{0.8060 $\pm$ 0.10} & $0.8056 \pm 0.07$ \\
    25 & \textbf{0.8200 $\pm$ 0.13} & $0.8130 \pm 0.05$ & $0.8185 \pm 0.18$ \\
    30 & \textbf{0.8272 $\pm$ 0.14} & $0.8244 \pm 0.07$ & $0.8250 \pm 0.14$ \\
    35 & $0.8328 \pm 0.19$ & $0.8307 \pm 0.13$ & \textbf{0.8328 $\pm$ 0.10} \\
    40 & \textbf{0.8391 $\pm$ 0.15} & $0.8377 \pm 0.09$ & $0.8378 \pm 0.07$ \\
    45 & $0.8411 \pm 0.27$ & \textbf{0.8426 $\pm$ 0.22} & $0.8410 \pm 0.08$ \\
    50 & $0.8413 \pm 0.16$ & \textbf{0.8471 $\pm$ 0.07} & $0.8447 \pm 0.19$ \\
    55 & \textbf{0.8483 $\pm$ 0.06} & $0.8476 \pm 0.09$ & $0.8462 \pm 0.14$ \\
    60 & $0.8492 \pm 0.08$ & \textbf{0.8500 $\pm$ 0.02} & $0.8488 \pm 0.12$ \\
    \bottomrule
    \bottomrule \\
  \end{tabular}
    \caption{The running time of the 3 strategies used for efficient incorporation of multiple latency targets. The design space used here is MobileNetV3}
\end{table}

\paragraph{Generalizing to other Design Spaces}
In addition to demonstrating the running time performance gains and the similarity in accuracy of the subnetworks, we also wanted to ensure that our strategies were compatible with any design space used for the OFA network. The other two design spaces OFA uses other than MobileNetV3 are Resnet50D and ProxylessNAS respectively. We run the exact set of experiments that we ran using MobileNetV3 with Resnet50D and ProxylessNAS. The results say the same story: we see the same trend with the running time as both design spaces show a significant difference between the vanilla method and the two strategies when there are multiple latency targets considered. As for the accuracy, both design spaces show no significant difference between the accuracies of the subnetworks obtained using either of the 3 strategies. 
\begin{figure}[!h]
    \centering
    \begin{subfigure}[b]{0.45\textwidth}
        \centering
        \includegraphics[width=\textwidth]{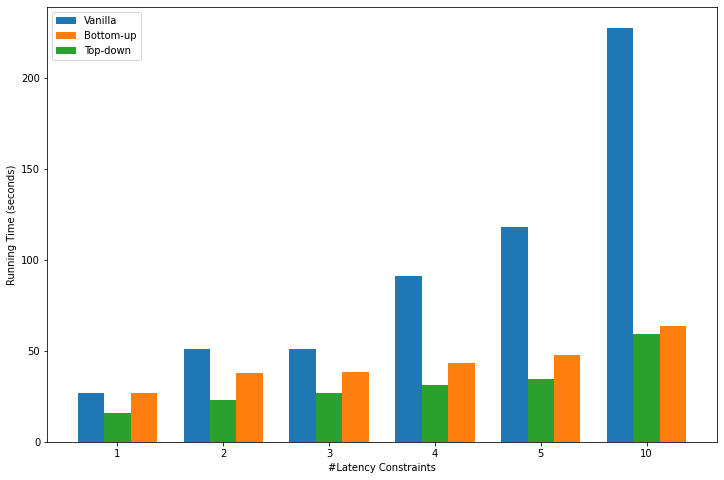}
        \caption{Resnet50D}
        \label{fig:y equals x}
    \end{subfigure}
    \hfill
    \begin{subfigure}[b]{0.45\textwidth}
        \centering
        \includegraphics[width=\textwidth]{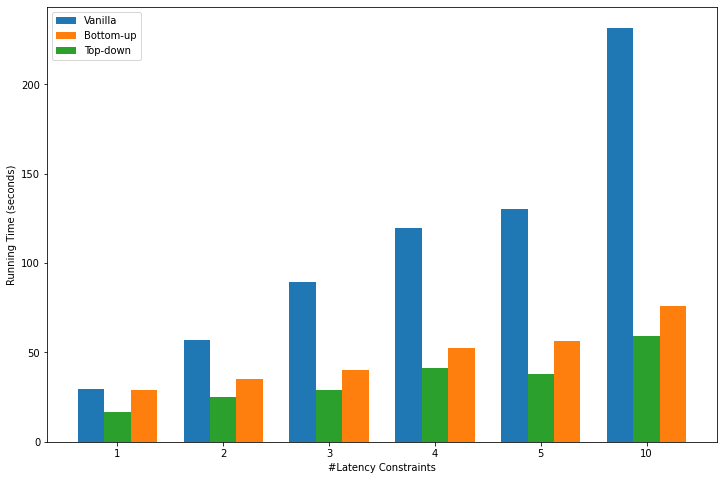}
        \caption{ProxylessNAS}
        \label{fig:three sin x}
    \end{subfigure}
    \hfill
    \caption{The running time of the 3 strategies used for efficient incorporation of multiple latency targets for the other two design spaces.}
    \label{fig:three graphs}
\end{figure}
\begin{figure}[!h]
    \centering
    \begin{subfigure}[b]{0.45\textwidth}
        \centering
        \includegraphics[width=\textwidth]{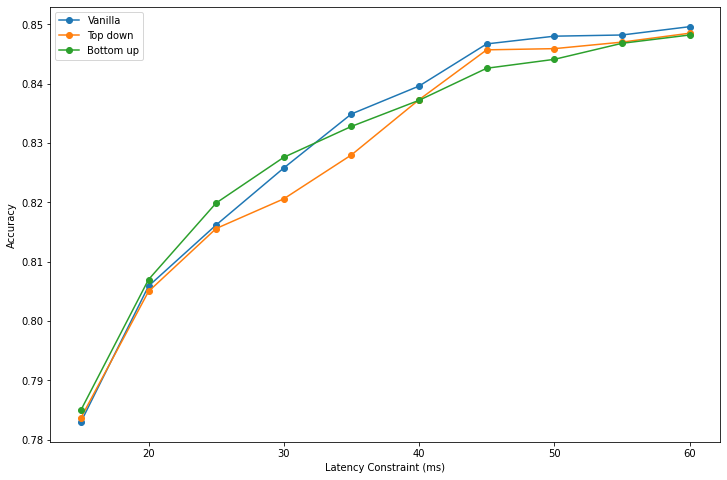}
        \caption{Resnet50D}
        \label{fig:y equals x}
    \end{subfigure}
    \hfill
    \begin{subfigure}[b]{0.45\textwidth}
        \centering
        \includegraphics[width=\textwidth]{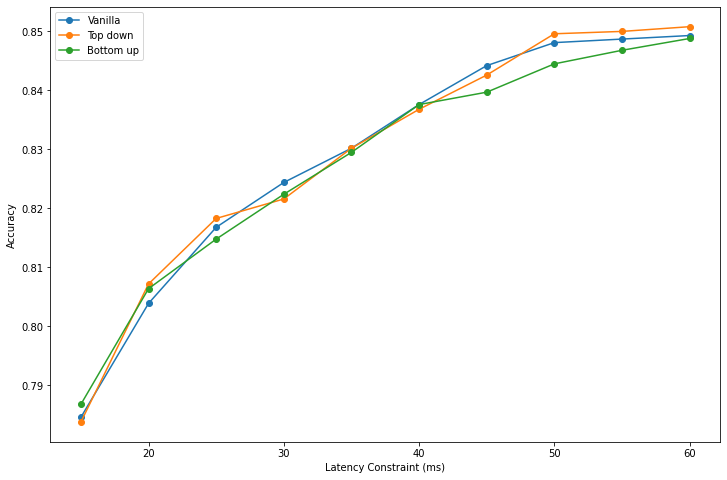}
        \caption{ProxylessNAS}
        \label{fig:three sin x}
    \end{subfigure}
    \hfill
    \caption{The validation accuracy of the optimal subnetworks found for each latency constraint by the 3 strategies for the other two design spaces.}
    \label{fig:three graphs}
\end{figure}

\section{Broader Impact}
The techniques we propose would contribute towards increasing the usability of techniques for finding optimal subnetworks for different latency constraints. Incorporating these techniques into the current OFA implementation would open up the usage of OFA to a wide variety of applications that would require running DNNs on a diverse set of target hardware. These techniques are often used in front-facing applications where high responsiveness improves the user experience. Improvements such as those in this paper would contribute to the area of improving the optimality of applications which use ML techniques. For example, it could improve the effectiveness of mobile applications that rely on automatic translation or voice/visual recognition.

\section{Conclusion}
Once-For-All (OFA) has proven to be an extremely effective technique in finding networks that are not only accurate, but also meet different latency constraints for a diverse set of target hardware. In this paper, we introduced two new strategies for improving OFA's incorporation of multiple latency constraints during the search phase. We demonstrated that these two strategies achieved significant running time performance gains over OFA's current implementation while not having to sacrifice the accuracy of the discovered subnetworks. We also showed that these strategies generalize to any design space used by OFA. We hope to continue working on these strategies and use the techniques developed in this paper to further latency-aware NAS algorithms in general.

\end{document}